\documentclass{article}
\usepackage{spconf,amsmath,graphicx,hyperref}
\usepackage{amsmath, amssymb}  
\usepackage{booktabs}  
\usepackage{multirow} 
\usepackage{graphicx}
\usepackage{tabularx}
\usepackage{pifont}

\usepackage{array}

\usepackage{booktabs}
\let\checkmark\relax
\usepackage{dingbat}

\title{UAUTrack: Towards Unified Multimodal Anti-UAV Visual Tracking}
\name{Qionglin Ren$^{1}$, Dawei Zhang$^{2,*}$, Chunxu Tian$^{1}$, Dan Zhang$^{3,*}$\thanks{* Corresponding author}}
\address{$^{1}$ College of Intelligent Robotics and Advanced Manufacturing, Fudan University, Shanghai, China\\
         $^{2}$ School of Computer Science and Technology, Zhejiang Normal University, Jinhua, China\\
         $^{3}$ Department of Mechanical Engineering, The Hong Kong Polytechnic University, Hong Kong, China}

\begin{document}
\ninept
\maketitle
\begin{abstract}
Research in Anti-UAV (Unmanned Aerial Vehicle) tracking has explored various modalities, including RGB, TIR, and RGB-T fusion. However, a unified framework for cross-modal collaboration is still lacking. Existing approaches have primarily focused on independent models for individual tasks, often overlooking the potential for cross-modal information sharing. Furthermore, Anti-UAV tracking techniques are still in their infancy, with current solutions struggling to achieve effective multimodal data fusion. To address these challenges, we propose UAUTrack, a unified single-target tracking framework built upon a single-stream, single-stage, end-to-end architecture that effectively integrates multiple modalities. UAUTrack introduces a key component: a text prior prompt strategy that directs the model to focus on UAVs across various scenarios. Experimental results show that UAUTrack achieves state-of-the-art performance on the Anti-UAV and DUT Anti-UAV datasets, and maintains a favourable trade-off between accuracy and speed on the Anti-UAV410 dataset, demonstrating both high accuracy and practical efficiency across diverse Anti-UAV scenarios. 
\end{abstract}
\begin{keywords}
Anti-UAV, Multimodal, Single Object Tracking, Text Prompt, Machine Learning
\end{keywords}
\section{Introduction}
In recent years, UAVs have rapidly developed, with their illegal use increasingly threatening public safety and transportation infrastructure. In response, Anti-UAV tracking has become a critical research area focused on tracking unauthorized aerial targets. This is a specific case of single-object visual tracking, aiming to continuously localize a target across video frames given its initial bounding box. While it shares the core formulation of general visual object tracking (VOT), Anti-UAV tracking presents additional challenges, such as small object size, rapid motion, and cluttered or low-visibility environments, often causing conventional trackers to fail. To address these issues, many approaches have been tailored to Anti-UAV scenarios \cite{ref2}. Moreover, available datasets cover multiple sensing modalities, including RGB, infrared (IR), and RGB-T fusion, as single-modality tracking struggles in adverse conditions, highlighting the need for more robust multimodal tracking frameworks.

\begin{figure}[htbp]
    \centering
    \includegraphics[width=\columnwidth]{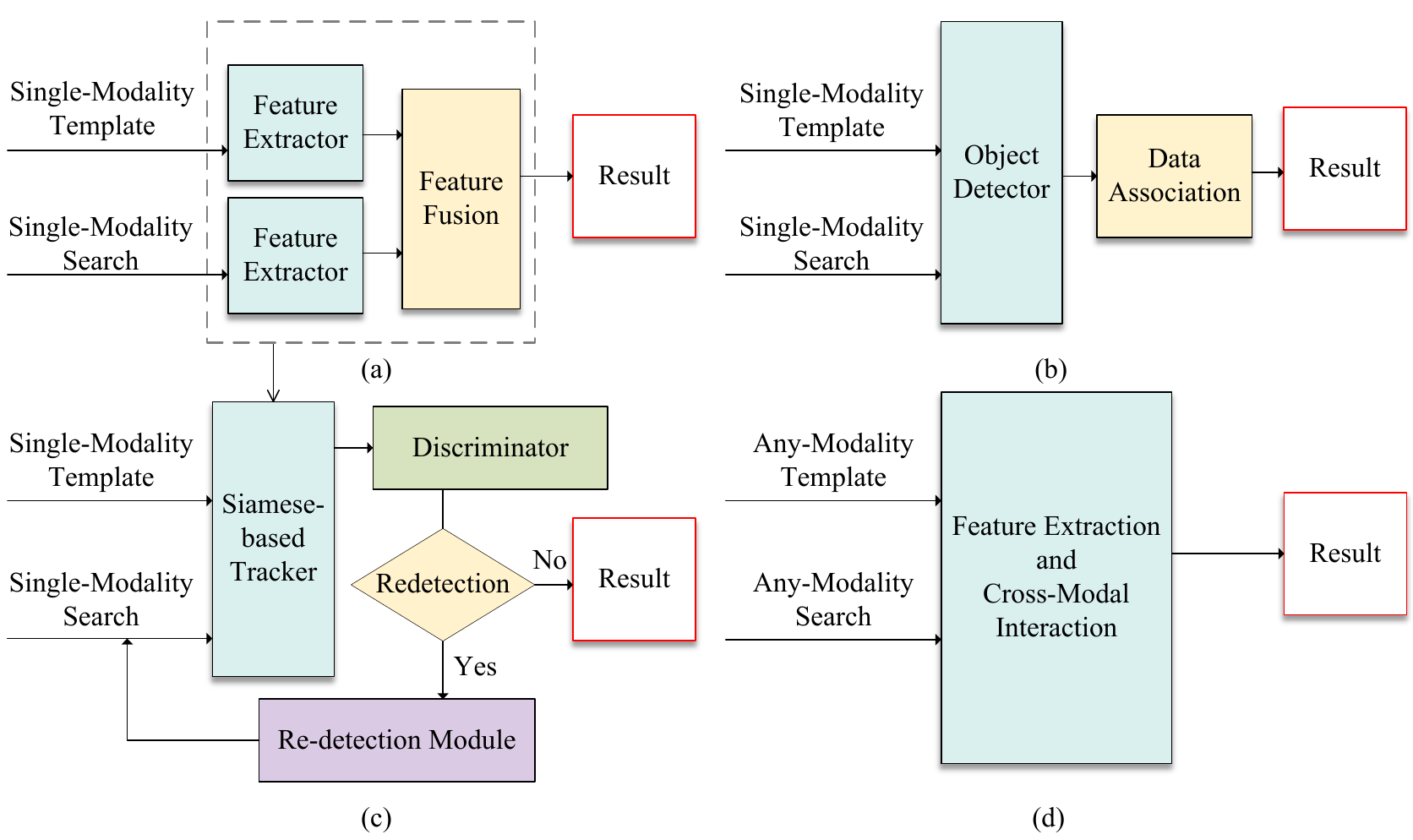} 
    \caption{Differences between our Anti-UAV tracking approach and previous methods. (a) Siamese-based trackers. (b) Detection-based trackers. (c) Hybrid trackers combining detection and Siamese networks. (d) Our unified end-to-end one-stream method. Any-Modality refers to the RGB, TIR, and RGB-T modalities.}
    \label{fig1}
\end{figure}

Current Anti-UAV tracking strategies primarily use Siamese structures or re-detection frameworks. As shown in Figure \ref{fig1}, the Siamese-based tracking diagram (a) \cite{li2022dual}, detection-based tracking diagram (b) \cite{wu2024biological}, and hybrid tracking diagram (c) \cite{cheng2022anti} each perform feature extraction and fusion separately, and the introduction of re-detection models adds to the computational burden. In contrast, our method (d) introduces a unified architecture that jointly extracts and fuses features across modalities in a single stream, eliminating the need for re-detection. This is the first framework in the Anti-UAV domain to unify tracking and fusion across modalities. By leveraging multimodal information holistically, it addresses various challenges while maintaining robust and efficient tracking in cluttered environments.

Most existing anti-UAV tracking methods rely on deep learning frameworks for infrared and visible-light tracking, often combining traditional trackers with custom strategies or using Siamese networks. In contrast, we propose UAUTrack, a unified multimodal UAV tracking framework that integrates RGB, thermal infrared (TIR), and RGB-T fusion modalities. By leveraging a Text Prior Prompt (TPP) module, UAUTrack processes data across modalities in a single stream, eliminating the need for separate architectures and enabling more effective fusion to enhance tracking performance.

Based on the above analysis, we propose UAUTrack, a multimodal Anti-UAV tracking framework capable of handling both single-modal and RGB-T video inputs. 

Overall, this study presents three main contributions:
\begin{itemize}
    \item We propose a modality-adaptive Anti-UAV tracking framework named UAUTrack, which unifies the tracking modalities in the existing Anti-UAV dataset and effectively leverages multi-modal information for UAV tracking.
    \item We design a Text Prior Prompt (TPP) strategy to provide semantic guidance, enabling accurate tracking without relying on an external detector. 
    \item We achieve state-of-the-art performance on widely used tracking benchmarks, including Anti-UAV and DUT AntiUAV, and demonstrate a favorable balance between tracking accuracy and speed on the Anti-UAV410 dataset.
\end{itemize}

\section{Method}

\begin{figure}[t]
    \centering
    \includegraphics[width=0.48\textwidth]{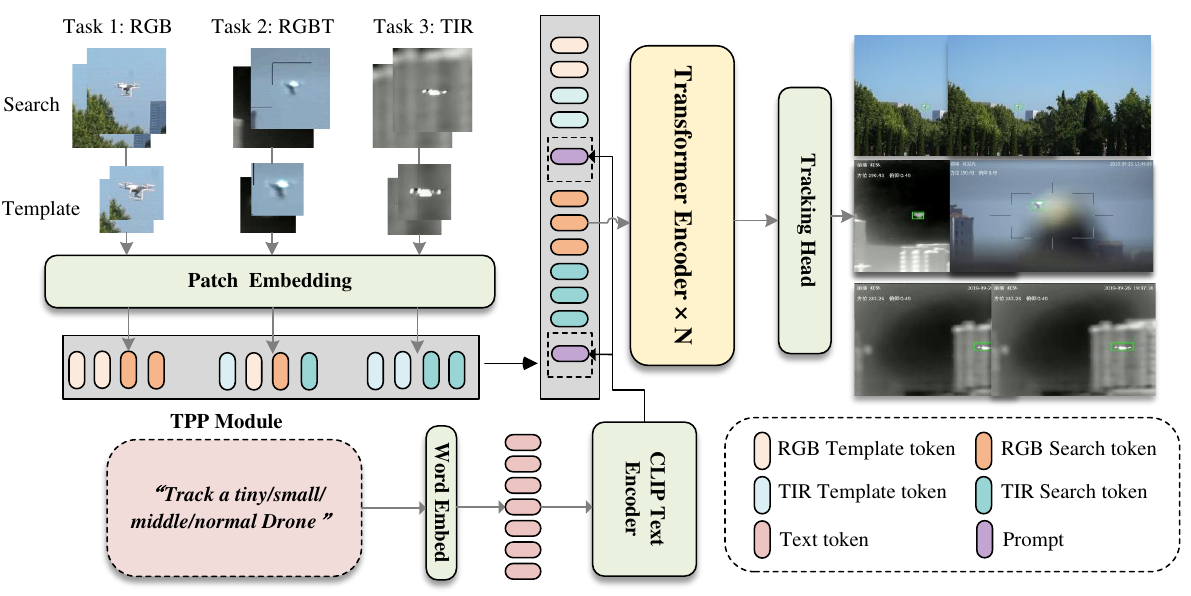} 
    \caption{Overall architecture of UAUTrack. A unified token embedding is employed to represent different modalities, including visible, thermal, and RGB-Thermal inputs. Text prior prompt (TPP) strategy are generated and fed into the transformer encoder, where they are fused with search and template tokens from each modality.}
    \label{fig:cross}
    \label{fig2}
\end{figure}

In this section, the proposed UAUTrack will be introduced in detail.  The overall architecture of UAUTrack is shown in Figure \ref{fig2}. Inspired by the multi-modal tracker SUTrack \cite{chen2025sutrack}, our backbone is built upon Fast-iTPN \cite{fastitpn}.

\subsection{Backbone}

The existing datasets in the Anti-UAV tracking domain include three modalities: visible (RGB), thermal infrared (TIR), and RGB-Thermal (RGB-T) fusion. To enable unified processing of these multimodal inputs, we preprocess the data into a consistent format. Specifically, RGB modality data \( I_{\text{RGB}} \in \mathbb{R}^{H \times W \times 3} \) and pseudo-colored TIR data \( I_{\text{TIR}} \in \mathbb{R}^{H \times W \times 3} \) are retained as individual three-channel inputs, while RGB-T inputs are obtained by concatenating the RGB and TIR images along the channel dimension, resulting in a six-channel representation \( I_{\text{RT}} \in \mathbb{R}^{H \times W \times 6} \). This preprocessing step yields a unified input structure \( I_U \in \mathbb{R}^{H \times W \times 6} \), allowing consistent feature processing across modalities. Each template and search image is then divided into non-overlapping patches, flattened into sequences, and linearly projected into \(D\)-dimensional embeddings, with positional and token type embeddings added. The embedded template vectors are denoted as \(V_{\text{t}}^{\text{RGB}}\) and \(V_{\text{t}}^{\text{TIR}}\), while the corresponding search embeddings are \(V_{\text{s}}^{\text{RGB}}\) and \(V_{\text{s}}^{\text{TIR}}\).

A self-attention mechanism is applied independently for the RGB-only and TIR-only tracking tasks, with the output denoted as \(A_S\), as defined in Equation (\ref{eq:unified_attention}). For the RGB-T tracking task, a cross-modal attention mechanism is employed to facilitate interaction between heterogeneous features from different modalities, producing the output \(A_C\). To fully exploit multimodal dependencies, the attention mechanism is further extended to jointly model inter-modal interactions, as detailed in Equation (\ref{eq1}).
\begin{align}
A_C &= \mathrm{Softmax} \left( 
    \frac{
        \left[ \mathbf{Q}_t^T ;\ \mathbf{Q}_s^T \right]
        \left[ \mathbf{K}_t^R ;\ \mathbf{K}_s^R \right]^\top
    }{
        \sqrt{d_k}
    }
\right) \cdot \left[ V_t^R ;\ V_s^R \right] \notag \\
&=
\left[ 
    \mathbf{S}_{tt}^{TR} \mathbf{V}_t^R + \mathbf{S}_{ts}^{TR} \mathbf{V}_s^R;\ 
    \mathbf{S}_{st}^{TR} \mathbf{V}_t^R + \mathbf{S}_{ss}^{TR} \mathbf{V}_s^R 
\right],
\label{eq1}
\end{align}
where $\mathbf{Q}$, $\mathbf{K}$, $\mathbf{V}$ are query, key, and value matrices respectively. \(\mathbf{S}\) is a measure of the similarity between RGB and TIR modal regions, which integrates the features of RGB and IR modalities in search and template images.
\noindent For the TIR or RGB single modality,  the self-attention mechanism is formulated as follows: 
\begin{equation}
A_{S}^{M} = \text{Softmax}\left( 
\frac{\left[ \mathbf{Q}_{t}^{M};\mathbf{Q}_{s}^{M} \right] 
\left( \left[ \mathbf{K}_{t}^{M};\mathbf{K}_{s}^{M} \right] \right)^{\top}}
{\sqrt{d_{k}}} 
\right) \cdot \left[ V_{t}^{M};V_{s}^{M} \right],
\label{eq:unified_attention}
\end{equation}

\noindent Here, $M \in \{T, R\}$ denotes the modality, where $T$ represents the TIR modality and $R$ represents the RGB modality. The variables $\mathbf{Q}^{M}, \mathbf{K}^{M}, V^{M}$ correspond to the Query, Key, and Value of modality $M$.

All resulting embeddings are concatenated along the sequence dimension and then fed into a stack of \(N\) encoder layers. The final backbone representation is defined recursively as:

\begin{equation}
    H_{ts}^{n} = E^{n-1}\left( H_{ts}^{n-1} \right), \quad n=1,2,\ldots,N,
\label{eq5}
\end{equation}
\noindent where \( H^{n}_{ts} \) denotes the output representation at the \( n \)-th encoder layer, encompassing both RGB and TIR features from the template and search branches. \( E^{n-1}(\cdot) \) represents the \((n{-}1)\)-th encoder layer. \( N \) is the total number of encoder layers, and the initial input \( H^{0}_{ts} \) corresponds to the concatenated embedding sequence of all modalities.
\subsection{Text Prior Prompt}
Given that the tracking target belongs to a fixed category, namely unmanned aerial vehicles (UAVs), we incorporate descriptive text based on the target's size as auxiliary visual information during training. Specifically, we use template-based textual prompts to provide additional supervision. These prompts take the form of ``track a \(G\) drone", where \(G =\{tiny,small,medium,normal\}\)  represents the UAV's size category. The size of the drone is determined by the diagonal length of its bounding box, following the partition scheme introduced in the Anti-UAV410 benchmark \cite{antiuav410} in detail. The text generation strategy is that the first frame comes from real values, followed by estimated values from each frame

We adopt a language model, CLIP-L \cite{clip}, as the text encoder. Given a textual description, the sentence is first tokenized and projected into a word embedding space, resulting in an initial embedding sequence $H_{L}^{0} \in \mathbb{R}^{{N_{K}} \times D}$, where $N_K$ is the maximum language length and \(D\) denotes the embedding dimension. The embedded sequence is then passed through a stack of 
\(K\) text encoder layers:
\begin{equation}
    H_{L}^{i}={{L}_{i}}\left( E_{L}^{i-1} \right),\text{  }i=1,2,\cdots ,K,
\label{eq6}
\end{equation}
A linear projection followed by a ReLU activation is applied to adjust the output dimension and extract a single token-level feature embedding. The transformed prompt is defined as:
\begin{equation}
    p_{L}^{i-1} = \text{Linear}\left( \operatorname{ReLU}\left( H_{L}^{i} \right) \right), \quad i = 1, 2, \dots, K ,
\end{equation}
\noindent where \(p_{L}^{i-1}\) denotes the processed language prompt at stage \({i-1}\). Then, the input prompt is split into two halves: $p_{L\_t}^{i-1}\in {\mathbb{R}^{n/2\times D}}$, for the template branch and $p_{L\_s}^{i-1}\in {\mathbb{R}^{n/2\times D}}$ for the search branch.
Finally, these language-guided prompts are integrated with the visual embeddings and passed through the unified transformer encoder. The fused representation at the $i_{th}$ encoder layer is computed as:
\begin{equation}
    H_{F}^{i}=E^{i-1}\left( \left[ p_{L\_t}^{i-1};H_{t}^{i-1};p_{L\_s}^{i-1};H_{s}^{i-1} \right] \right),\quad i=1,2,\cdots ,N,
\label{eq8}
\end{equation}
\noindent where \(E^{i-1}\) denotes the $i_{th}$ unified transfomer encoder layer.

\subsection{Head and Loss}

For training, weighted focal loss\cite{weighted_focal_loss} is used for classification, while a combination of L1 loss and Generalized IoU (GIoU) is applied for bounding box regression. The overall loss function is defined as:
\begin{equation}
    \mathcal{L}={{\lambda }_{G}}{{\mathcal{L}}_{\text{class }}}+{{\lambda }_{giou}}{{\mathcal{L}}_{giou\text{ }}}+{{\lambda }_{{{L}_{1}}}}{{\mathcal{L}}_{{{L}_{1}}}}+{{\lambda }_{task}}{{\mathcal{L}}_{\text{task }}},
\label{eq12}
\end{equation}
\noindent the corresponding parameters are set to 1, 2, 5, and 1, respectively.

\section{Experiments}


\subsection{Implementation Details}
UAUTrack is implemented using PyTorch 2.1 and trained on two NVIDIA RTX 6000 ADA GPUs with batch size 32. SUTrack \cite{chen2025sutrack} is adopted as the baseline, and its pre-trained model (trained for 180 epochs) is used for initialization.

The fine-tuning process is performed using three datasets: Anti-UAV (RGB), Anti-UAV410 (Infrared), and Anti-UAV RGB-T. It is conducted for 20 epochs, with each epoch comprising 60,000 samples. The search and template region sizes are set to 224×224 and 112×112, respectively. The model is optimized using the AdamW optimizer with an initial learning rate of $1 \times 10^{-4}$ and a weight decay of $1 \times 10^{-4}$. Fixed parameters are initialized from the base model following the method described in \cite{chen2025sutrack}. Among several fine-tuning strategies evaluated, full fine-tuning consistently yields the best performance, as further confirmed by our ablation studies.

In the proposed online template updating strategy, an update is performed only when two conditions are simultaneously satisfied: (1) the current frame index is an integer multiple of the predefined update interval (set to 25), and (2) the tracking confidence score of the current frame exceeds a reliability threshold of 0.7. To further stabilize the updating process and suppress abrupt changes, a Hanning window is applied as a spatial penalty during the update.
\subsection{Evaluation Metrics}
In the context of Anti-UAV tracking, in addition to standard evaluation metrics such as Success (AUC), Precision (P), and Normalized Precision (\( \mathrm{P}_{Norm} \)), the State Accuracy (SA) is also widely adopted \cite{antiuav410}. The SA metric is  defined as:
\begin{equation}
SA=\frac{1}{T}\sum\limits_{t=1}^{T}{\left( Io{{U}_{t}}\times \delta \left( {{v}_{t}}>0 \right)+{{p}_{t}}\times \left( 1-\delta \left( {{v}_{t}}>0 \right) \right) \right)},
\label{eq12}
\end{equation}
\noindent where \(T\) represents the total number of samples, and \(t\) represents a single sample. The term \( IoU_t \) measures the Intersection-over-Union between the predicted bounding box and the ground truth at time \( t \). The visibility flag \( v_t \) indicates whether the target is visible; the indicator function \( \delta(v_t > 0) \) returns 1 when the target is visible and 0 otherwise. The presence variable \( p_t \) represents the prediction status, taking value 0 when a detection is present and 1 when it is missing.
\subsection{Qualitative analysis with state-of-the-art methods}
\begin{figure}[htbp]
    \centering
    \includegraphics[width=0.48\textwidth]{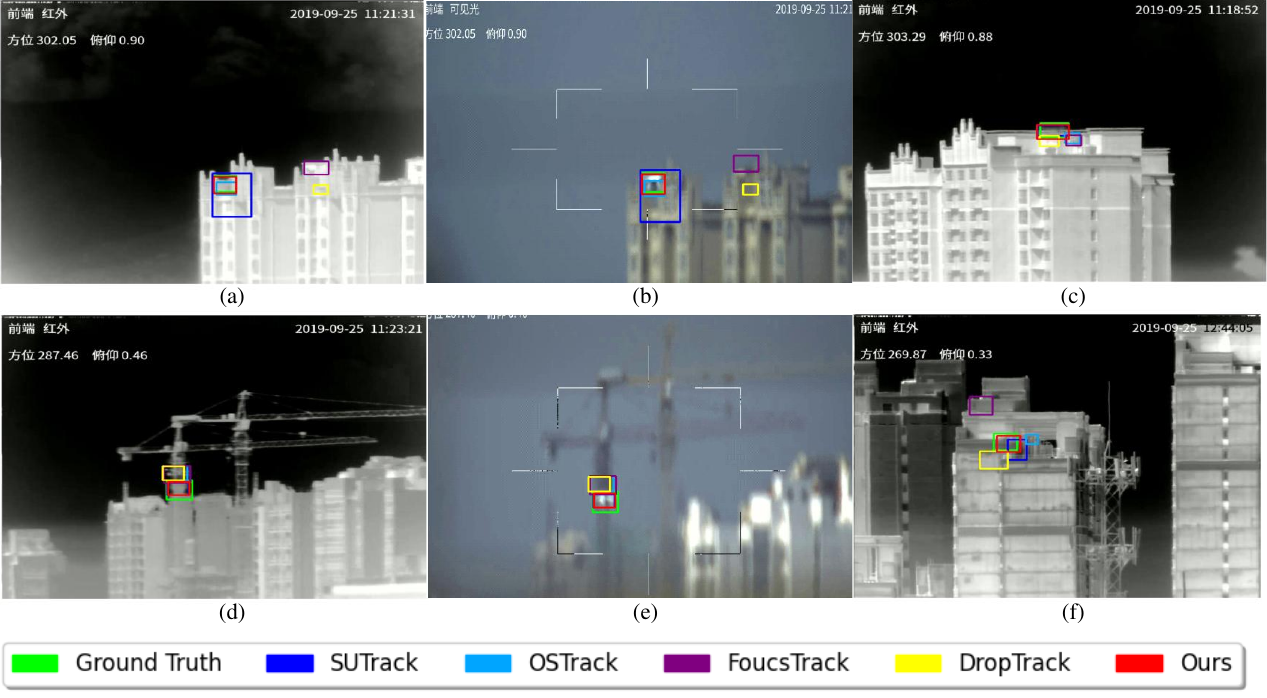} 
    \caption{Qualitative visualization of common challenges in Anti-UAV tracking, including (a)TIR Thermal Crossover (TC), (b)RGB TC, (c)TIR Scale Variation (SV), (d)TIR Low Resolution (LR), (e)RGB LR, (f)TIR Fast Motion (FM). }
    \label{fig3}
\end{figure}

For qualitative analysis, we visualize representative challenges in the anti-UAV infrared benchmark, including Thermal Crossover (TC), Scale Variation (SV), Low Resolution (LR), and Fast Motion (FM). As shown in Figure \ref{fig3}, other single-modality trackers tend to lose the target or experience bounding box drift under these conditions, whereas UAUTrack consistently maintains accurate focus on the target, demonstrating the robustness and generalization capability of the proposed framework.
\subsection{Quantitative analysis with state-of-the-art methods}
In this section, we compare our proposed UAUTrack with other state-of-the-art (SOTA) trackers on three benchmark datasets: Anti-UAV410 \cite{antiuav410}, Anti-UAV \cite{antiuav}, and DUT Anti-UAV \cite{dutantiuav}.


Anti-UAV is a multimodal anti-drone dataset that contains both RGB and TIR modalities. As shown in Table \ref{table1}, under the TIR single-modality setting, UAUTrack achieves state-of-the-art performance, surpassing the second-best single-modality tracker by 1.1\%, 0.6\%, and 3.0\% in AUC, normalized precision, and SA, respectively. Furthermore, as shown in Table \ref{table2}, UAUTrack outperforms SUTrack by 1.6\% overall on the RGB modality. The results in Tables \ref{table1} and \ref{table2} validate that, compared to single-modality trackers, the superior performance of UAUTrack demonstrates the effectiveness of multimodal fusion in enhancing tracking accuracy. In addition, compared with multimodal trackers such as SUTrack, our use of textual prompt further improves overall tracking performance and speed.

\begin{table*}[htbp]
\centering
\caption{Evaluation results on Anti-UAV, Anti-UAV410 and DUT Anti-UAV datasets.}
\begin{tabular}{l|cccc|cccc|ccc|c}
\toprule
\multirow{2}{*}{Method} 
& \multicolumn{4}{c|}{Anti-UAV (TIR)} 
& \multicolumn{4}{c|}{Anti-UAV410} 
& \multicolumn{3}{c|}{DUT Anti-UAV} 
& \multirow{2}{*}{Fps} \\
\cmidrule(lr){2-5} \cmidrule(lr){6-9} \cmidrule(lr){10-12}
& AUC & P & \( \mathrm{P}_{Norm} \) & SA 
& AUC & P & \( \mathrm{P}_{Norm} \) & SA 
& AUC & P & \( \mathrm{P}_{Norm} \) 
& \\
\midrule
DiMP \cite{dimp}                  & 52.5 & 58.3 & 55.4 & -- & -- & -- & -- & -- & 57.8 & 83.1 & 75.6 & 29 \\
TransT \cite{transt}              & 58.5 & 44.3 & 51.4 & -- & -- & -- & -- & -- & 58.6 & 83.2 & 76.5 & 26 \\
LTMU \cite{LTMU}                  & 60.0 & 57.9 & 59.0 & -- & -- & -- & -- & -- & 60.8 & 85.8 & 78.3 & 16 \\
SiamFusion \cite{siamfusion}      & 66.1 & 60.7 & 63.4 & -- & -- & -- & -- & -- & --   & --   & --   & 14 \\
OSTrack \cite{ostrack}            & 59.2 & 79.4 & 77.5 & 60.2 & 53.7 & 73.9 & 70.9 & 54.7 & 63.1 & 86.8 & 88.1 & 137 \\
ROMTrack \cite{romtrack}          & 59.4 & 78.9 & 59.4 & 60.5 & 54.7 & 74.5 & 71.7 & 55.7 & --   & --   & --   & 8 \\
ZoomTrack \cite{kou2023zoomtrack} & 63.5 & 86.0 & 83.4 & 64.5 & 58.4 & 81.2 & 77.4 & 59.4 & --   & --   & --   & 154 \\
DropTrack \cite{droptrack}        & 64.2 & 85.8 & 83.2 & 65.2 & 59.2 & 82.2 & 78.2 & 60.2 & 64.2 & 87.0 & 85.1 & 151 \\
SUTrack \cite{chen2025sutrack}    & 58.7 & 77.4 & 76.0 & 61.0 & 55.2 & 74.0 & 72.2 & 56.7 & --   & --   & --   & 33 \\
GlobalTrack \cite{huang2020globaltrack} & 64.2 & 88.9 & -- & -- & 65.1 & 85.3 & -- & 66.3 & --   & --   & --   & 9 \\
FocusTrack \cite{wang2025focustrack}   & 67.7 & 90.9 & 88.4 & 68.9 & 62.6 & 86.7 & 83.0 & 63.9 & --   & --   & --   & 44 \\
SiamDT \cite{antiuav410}          & -- & -- & -- & -- & 66.8 & 87.6 & -- & 68.2 & --   & --   & --   & 8 \\
ARTrackV2 \cite{bai2024artrackv2} & -- & -- & -- & -- & 51.5 & 70.4 & -- & 52.5 & --   & --   & --   & 30 \\
UAUTrack (Ours)                   & 68.8 & 89.7 & 89.0 & 71.9 & 64.2 & 85.0 & 82.9 &66.3 & 65.8 &87.3& 91.8 & 45 \\
\bottomrule
\end{tabular}
\label{table1}
\end{table*}

\begin{table}[t]
\centering
\caption{Evaluation results (SA) on Anti-UAV dataset.}
\begin{tabular}{lccc}
\toprule
Method & TIR & RGB & Total \\
\midrule
DiMP \cite{dimp}        & 52.5 & 58.3 & 55.4 \\
SiamRPN++ \cite{li2019siamrpn++}  & 45.0 & 49.7 & 47.4 \\
LTMU \cite{LTMU}        & 60.0 & 57.9 & 59.0 \\
TransT \cite{transt}      & 58.5 & 44.3 & 51.4 \\
KeepTrack \cite{keeptrack}   & 56.8 & 59.4 & 58.1 \\
DFSC \cite{antiuav}        & 66.0 & 69.8 & 67.9 \\
SiamFusion \cite{siamfusion} & 66.1 & 60.7 & 63.4 \\
SUTarck \cite{chen2025sutrack}     & 70.3 & 65.9 & 62.3 \\
UAUTrack (Ours)   & \textbf{71.9} & \textbf{76.2} & \textbf{74.0} \\
\bottomrule
\end{tabular}
\label{table2}
\end{table}

Anti-UAV410 is a single-modality infrared anti-drone dataset comprising 410 sequences.
As shown in Table \ref{table1}, UAUTrack demonstrates outstanding performance in both Precision and State Accuracy (SA) across most competing trackers. Compared to the global tracker SiamDT \cite{antiuav410}, which achieves a peak SA of 68.2\%, UAUTrack exhibits superior efficiency, achieving nearly a 6× speedup while maintaining competitive accuracy, with only a 1.9\% drop in SA. While SiamDT adopts a global tracking paradigm, UAUTrack is based on a local tracking strategy that is significantly more resource-efficient and computationally lightweight. This design enables UAUTrack to effectively balance tracking speed and accuracy, making it well-suited for real-time anti-drone applications. Compared to GlobalTrack, UAUTrack achieves similar SA scores but is five times faster, further demonstrating the advantages of our framework.

DUT Anti-UAV is an RGB single-modality anti-drone dataset that includes both short-term and long-term tracking scenarios. As shown in Table \ref{table1}, UAUTrack outperforms the second-best tracker by 1.6 \% in AUC and 0.3\% in Precision.

\subsection{Ablation Study}


To evaluate the effectiveness of our training strategy and core components, we conducted a comprehensive ablation study on the Anti-UAV dataset. A series of experiments were performed to assess various fine-tuning strategies, while also considering the introduction of the Text Prior Prompt (TPP) strategy.

The experimental configurations are as follows: (1) Baseline: no fine-tuning; (2) Frozen Encoder: fine-tuning with the encoder frozen and the decoder trainable; (3) LoRA Fine-tuning: lightweight parameter adaptation using LoRA; (4) Adapter Fine-tuning: simple attention-based adapter for fine-tuning; (5) Full Fine-tuning: all parameters unfrozen. These configurations do not incorporate our proposed strategy; (6) Fully fine tune the single-modality special model: only use the single-modality data for training.(7) Full Fine-tuning with TPP: full fine-tuning with the inclusion of the Text Prior Prompt strategy. Note that (1)-(5) configurations do not include TPP. 

As shown in Table ~\ref{table4}, fine-tuning with a frozen backbone and resource-efficient strategies such as LoRA exhibit limited performance gains. In contrast, the straightforward approach of full fine-tuning leads to a significant improvement in accuracy. This indicates that optimizing all model parameters enables more effective extraction of low-level features and better adaptation to the specific challenges of anti-UAV tracking scenarios. When further introducing the text guided prompt strategy, the accuracy will be increased by an additional 5\% compared to (7) and (5). These results indicate that text priors can effectively guide the tracking process, achieving higher accuracy and efficiency. At the same time, compared with (7) and (6), the accuracy of the multimodal model improved by 4.4\% compared to the dedicated single modal model, which also demonstrates the advantages of our framework. Note that (6) cannot be directly compared with (1) - (5) due to the control variable method.
\begin{table}[t]
\centering
\setlength{\tabcolsep}{6pt} 
\renewcommand{\arraystretch}{1.2} 

\caption{Fine-tuning strategy and component analysis on Anti-UAV dataset.}
\begin{tabular}{c|ccc|c|c}
\toprule
Method & AUC & Precision & Fps & TPP & Multimodal \\
\midrule
\ding{172} & 58.7 & 77.4 & 37 &  & \checkmark \\
\ding{173} & 61.6 & 80.1 & 47 &  & \checkmark \\
\ding{174} & 62.3 & 83.3 & 66 &  & \checkmark \\
\ding{175} & 62.9 & 83.6 & 32 &  & \checkmark \\
\ding{176} & 63.9 & 84.0 & 47 &  & \checkmark \\
\midrule  
\ding{177} & 64.4 & 85.4 & 46 & \checkmark &  \\
\ding{178} & \textbf{68.8} & \textbf{89.7} & 45 & \checkmark & \checkmark \\
\bottomrule
\end{tabular}
\label{table4}
\end{table}




\section{Conclusion}
In this work, we propose UAUTrack, a unified framework for anti-UAV tracking that integrates three representative tasks into a single model. Unlike prior methods, UAUTrack is the first to introduce a multi-modal architecture tailored for Anti-UAV scenarios, enabling cross-modality tracking through a unified training process. By incorporating a text prior prompt strategy, our framework achieves accurate tracking without relying on re-detection.  Extensive experiments on three mainstream Anti-UAV benchmarks have verified the effectiveness and versatility of our proposed architecture and modules. Future work will explore extending the framework to more complex real-world environments.

\newpage

\bibliographystyle{IEEEbib}
\bibliography{refs2}

\begin{thebibliography}{10}

\bibitem{ref2}
Guanghai Ding, Yihua Ren, Yuting Liu, Qijun Zhao, and Shuiwang Li,
\newblock ``Vision-based anti unmanned aerial technology: Opportunities and challenges,''
\newblock {\em arXiv:2507.10006}, 2025.

\bibitem{li2022dual}
Shaogang Li, Jin Gao, Liang Li, Gang Wang, Yizheng Wang, and Xin Yang,
\newblock ``Dual-branch approach for tracking uavs with the infrared and inverted infrared image,''
\newblock in {\em ICSP}, 2022, pp. 1803--1806.

\bibitem{wu2024biological}
Tongyan Wu, Haibin Duan, and Zhigang Zeng,
\newblock ``Biological eagle eye-based correlation filter learning for fast uav tracking,''
\newblock {\em IEEE Transactions on Instrumentation and Measurement}, 2024.

\bibitem{cheng2022anti}
Feng Cheng, Zhibo Liang, Gaoliang Peng, Shaohui Liu, Sijue Li, and Mengyu Ji,
\newblock ``An anti-uav long-term tracking method with hybrid attention mechanism and hierarchical discriminator,''
\newblock {\em Sensors}, vol. 22, no. 10, pp. 3701, 2022.

\bibitem{chen2025sutrack}
Xin Chen, Ben Kang, Wanting Geng, Jiawen Zhu, Yi~Liu, Dong Wang, and Huchuan Lu,
\newblock ``Sutrack: Towards simple and unified single object tracking,''
\newblock in {\em AAAI}, 2025, pp. 2239--2247.

\bibitem{fastitpn}
Yunjie Tian, Lingxi Xie, Jihao Qiu, Jianbin Jiao, Yaowei Wang, Qi~Tian, and Qixiang Ye,
\newblock ``Fast-itpn: Integrally pre-trained transformer pyramid network with token migration,''
\newblock {\em IEEE Transactions on Pattern Analysis and Machine Intelligence}, 2024.

\bibitem{antiuav410}
Bo~Huang, Jianan Li, Junjie Chen, Gang Wang, Jian Zhao, and Tingfa Xu,
\newblock ``Anti-uav410: A thermal infrared benchmark and customized scheme for tracking drones in the wild,''
\newblock {\em IEEE Transactions on Pattern Analysis and Machine Intelligence}, vol. 46, no. 5, pp. 2852--2865, 2023.

\bibitem{clip}
Alec Radford, Jong~Wook Kim, Chris Hallacy, Aditya Ramesh, Gabriel Goh, Sandhini Agarwal, Girish Sastry, Amanda Askell, Pamela Mishkin, Jack Clark, Gretchen Krueger, and Ilya Sutskever,
\newblock ``Learning transferable visual models from natural language supervision,''
\newblock in {\em ICML}, 2021, pp. 8748--8763.

\bibitem{weighted_focal_loss}
Hei Law and Jia Deng,
\newblock ``Cornernet: Detecting objects as paired keypoints,''
\newblock in {\em ECCV}, 2018.

\bibitem{antiuav}
Nan Jiang, Kuiran Wang, Xiaoke Peng, Xuehui Yu, Qiang Wang, Junliang Xing, Guorong Li, Guodong Guo, Qixiang Ye, and Jianbin Jiao,
\newblock ``Anti-uav: A large-scale benchmark for vision-based uav tracking,''
\newblock {\em IEEE Transactions on Multimedia}, vol. 25, pp. 486--500, 2021.

\bibitem{dutantiuav}
Jie Zhao, Jingshu Zhang, Dongdong Li, and Dong Wang,
\newblock ``Vision-based anti-uav detection and tracking,''
\newblock {\em IEEE Transactions on Intelligent Transportation Systems}, vol. 23, no. 12, pp. 25323--25334, 2022.

\bibitem{dimp}
Goutam Bhat, Martin Danelljan, Luc Van~Gool, and Radu Timofte,
\newblock ``Learning discriminative model prediction for tracking,''
\newblock in {\em ICCV}, 2019.

\bibitem{transt}
Xin Chen, Bin Yan, Jiawen Zhu, Dong Wang, Xiaoyun Yang, and Huchuan Lu,
\newblock ``Transformer tracking,''
\newblock in {\em CVPR}, 2021, pp. 8126--8135.

\bibitem{LTMU}
Kenan Dai, Yunhua Zhang, Dong Wang, Jianhua Li, Huchuan Lu, and Xiaoyun Yang,
\newblock ``High-performance long-term tracking with meta-updater,''
\newblock in {\em CVPR}, 2020.

\bibitem{siamfusion}
Zhihao Zhang, Lei Jin, Shengjie Li, JianQiang Xia, Jun Wang, Zun Li, Zheng Zhu, Wenhan Yang, PengFei Zhang, Jian Zhao, and Bo~Zhang,
\newblock ``Modality meets long-term tracker: A siamese dual fusion framework for tracking uav,''
\newblock in {\em ICIP}, 2023, pp. 1975--1979.

\bibitem{ostrack}
Botao Ye, Hong Chang, Bingpeng Ma, Shiguang Shan, and Xilin Chen,
\newblock ``Joint feature learning and relation modeling for tracking: A one-stream framework,''
\newblock in {\em ECCV}, 2022, pp. 341--357.

\bibitem{romtrack}
Yidong Cai, Jie Liu, Jie Tang, and Gangshan Wu,
\newblock ``Robust object modeling for visual tracking,''
\newblock in {\em ICCV}, 2023, pp. 9589--9600.

\bibitem{kou2023zoomtrack}
Yutong Kou, Jin Gao, Bing Li, Gang Wang, Weiming Hu, Yizheng Wang, and Liang Li,
\newblock ``Zoomtrack: Target-aware non-uniform resizing for efficient visual tracking,''
\newblock {\em NeurIPS}, vol. 36, pp. 50959--50977, 2023.

\bibitem{droptrack}
Qiangqiang Wu, Tianyu Yang, Ziquan Liu, Baoyuan Wu, Ying Shan, and Antoni~B. Chan,
\newblock ``Dropmae: Masked autoencoders with spatial-attention dropout for tracking tasks,''
\newblock in {\em CVPR}, 2023, pp. 14561--14571.

\bibitem{huang2020globaltrack}
Lianghua Huang, Xin Zhao, and Kaiqi Huang,
\newblock ``Globaltrack: A simple and strong baseline for long-term tracking,''
\newblock {\em AAAI}, vol. 34, no. 7, pp. 11037--11044, 2020.

\bibitem{wang2025focustrack}
Ying Wang, Tingfa Xu, and Jianan Li,
\newblock ``Focustrack: A self-adaptive local sampling algorithm for efficient anti-uav tracking,''
\newblock {\em IEEE Transactions on Geoscience and Remote Sensing}, 2025.

\bibitem{bai2024artrackv2}
Yifan Bai, Zeyang Zhao, Yihong Gong, and Xing Wei,
\newblock ``Artrackv2: Prompting autoregressive tracker where to look and how to describe,''
\newblock in {\em CVPR}, 2024, pp. 19048--19057.

\bibitem{li2019siamrpn++}
Bo~Li, Wei Wu, Qiang Wang, Fangyi Zhang, Junliang Xing, and Junjie Yan,
\newblock ``Siamrpn++: Evolution of siamese visual tracking with very deep networks,''
\newblock in {\em CVPR}, 2019.

\bibitem{keeptrack}
Christoph Mayer, Martin Danelljan, Danda~Pani Paudel, and Luc Van~Gool,
\newblock ``Learning target candidate association to keep track of what not to track,''
\newblock in {\em ICCV}, 2021, pp. 13444--13454.

\end{thebibliography}

\end{document}